\documentclass{article}

\usepackage{amsfonts}
\usepackage{amsmath}
\usepackage{subfig}
\usepackage[font=small]{caption}
\usepackage{tabularx}
\usepackage{bm}
\usepackage{multirow}
\usepackage{setspace}
\usepackage{color} 
\usepackage{float}
\usepackage[linesnumbered,ruled,vlined ]{algorithm2e}
\SetAlTitleFnt{\footnotesize}
\SetAlCapNameFnt{\footnotesize}
\SetAlCapFnt{\footnotesize}

\usepackage{microtype}
\usepackage{graphicx}
\usepackage{booktabs} 
\DeclareMathOperator*{\argmax}{arg\,max}
\DeclareMathOperator*{\argmin}{arg\,min}

\expandafter\def\expandafter\normalsize\expandafter{%
    \normalsize
    \setlength\abovedisplayskip{5pt}
    \setlength\belowdisplayskip{5pt}
    \setlength\abovedisplayshortskip{5pt}
    \setlength\belowdisplayshortskip{5pt}
}

\newtheorem{theorem}{Theorem}

\newcommand{\m}{\mathrm{max}}
\newcommand{\SO}{\mathrm{SO}}
\newcommand{\mS}{\mathrm{S}}

\usepackage{hyperref}

\usepackage[accepted]{icml2019}

\icmltitlerunning{Multi-Frequency Vector Diffusion Maps}

\begin{document}
\begin{spacing}{0.92}

\twocolumn[
\icmltitle{Multi-Frequency Vector Diffusion Maps}



\icmlsetsymbol{equal}{*}

\begin{icmlauthorlist}
\icmlauthor{Yifeng Fan}{uiuc}
\icmlauthor{Zhizhen Zhao}{uiuc}

\end{icmlauthorlist}

\icmlaffiliation{uiuc}{Department of Electrical and Computer Engeneering, Coordinated Science Laboratory, University of Illinois at Urbana-Champaign, Illinois, USA}

\icmlcorrespondingauthor{Yifeng Fan}{yifengf2@illinois.edu}

\icmlkeywords{Machine Learning, ICML}

\vskip 0.3in
]



\printAffiliationsAndNotice{}  

\begin{abstract}
We introduce multi-frequency vector diffusion maps (MFVDM), a new framework for organizing and analyzing high dimensional datasets. The new method is a mathematical and algorithmic generalization of vector diffusion maps (VDM) and other non-linear dimensionality reduction methods. 
MFVDM combines different nonlinear embeddings of the data points defined with multiple unitary irreducible representations of the alignment group that connect two nodes in the graph. 
We illustrate the efficacy of MFVDM on synthetic data generated according to a random graph model and cryo-electron microscopy image dataset. The new method achieves better nearest neighbor search and alignment estimation than the state-of-the-arts VDM and diffusion maps (DM) on extremely noisy data. 

\end{abstract}



\section{Introduction}
\label{sec:intro}
Nonlinear dimensionality reduction methods, such as locally linear embedding (LLE)~\cite{roweis2000nonlinear}, ISOMAP~\cite{tenenbaum2000global}, Hessian LLE~\cite{donoho2003hessian}, Laplacian eigenmaps~\cite{belkin2002laplacian,belkin2003laplacian}, and diffusion maps (DM)~\cite{coifman2006diffusion} are invaluable tools for embedding complex data in a low dimensional space and for regression problems on graphs and manifolds. 
To this end, those methods assume that the high-dimensional data lies on a low dimensional manifold and local affinities in a weighted neighborhood graph are used to learn the global structure of the data. 
Spectral clustering~\cite{nadler2006diffusion,von2007tutorial}, semi-supervised learning~\cite{zhu2006semi,goldberg2009multi,yang2016revisiting}, out-of-sample extension~\cite{belkin2006manifold}, image denoising~\cite{gong2010locally, singer2009diffusion} share  similar geometrical considerations.  Those techniques are either directly or indirectly related to the heat kernel for functions on the data. Vector diffusion maps (VDM)~\cite{SingerWu2012VDM} generalizes DM to define heat kernel for vector fields on the data manifold. The corresponding adjacency matrix is based on edge weights and orthogonal transformations between connected nodes. 
Using the spectral decomposition of the matrix, VDM defines a metric for the data to indicate the closeness of the data points on the manifold. For some applications, the vector diffusion metric is beneficial, since it takes into account linear transformations, and as a result, it provides a better organization of the data. 
However, for extremely noisy data, VDM nearest neighbor search may fail at identifying the true nearby points on the manifold. This results in shortcut edges that connect points with large geodesic distances on the manifold. 

To address this issue, we introduce a new algorithm called multi-frequency vector diffusion maps (MFVDM) to represent and organize complex high-dimensional data, exhibiting a non-trivial group invariance. To this end, we augment VDM 
 with multiple irreducible representations of the compact group 
to improve the 
rotationally invariant nearest neighbor search and the alignment estimation between nearest neighbor pairs, when the initial estimation contains a large number of outliers due to noise. 
Specifically, we define a set of kernels, denoted by $W_k$, using multiple irreducible representations of the compact alignment group indexed by integer $k$ and introduce the corresponding frequency-$k$-VDMs. The MFVDM is constructed by concatenating all the frequency-$k$-VDMs up to a cutoff $k_\m$. 
We use the new embeddings to identify nearest neighbors. The eigenvectors of the normalized $W_k$ are used to estimate the pairwise alignments between nearest neighbors. This framework also extends the mathematical theory of cryo-electron microscopy (EM) image analysis~\cite{singer2011viewing,hadani2011representation, giannakis2012symmetries, schwander2012symmetries, dashti2014trajectories}. 
We show that MFVDM outperforms VDM and DM for data sampled from low-dimensional manifolds, when a large proportion of the edge connections are corrupted. MFVDM is also able to improve the nearest neighbor search and rotational alignment for 2-D class averaging in cryo-EM.

\section{Preliminaries and Problem Setup}
\label{sec:setup}
Given a dataset $x_i \in \mathbb{R}^l$ for $i = 1, \dots, n$, we assume that the data lie on or close to a low dimensional smooth manifold $\mathcal{X}$ of intrinsic dimension $d \ll l$. Suppose that $\mathcal{G}$ is a compact Lie group, which has unitary irreducible representations according to Peter-Weyl theorem. The data space $\mathcal{X}$ is closed under $\mathcal{G}$ if for all $g \in \mathcal{G}$ and all $x \in \mathcal{X}$, $g \cdot x \in \mathcal{X}$, where `$\cdot$' denotes the group action. 
The $\mathcal{G}$-invariant distance between two data points is defined as,
\begin{equation}
    \label{eq:rid}
    d_{ij} = \min_{g \in \mathcal{G}} \| x_i - g\cdot x_j\|, 
\end{equation}
and the associated optimal alignment is,
\begin{equation}
     \label{eq:opt_g}
    g_{ij} = \argmin_{g \in \mathcal{G}} \| x_i - g\cdot x_j\|. 
\end{equation}
We assume that the optimal alignment is unique and construct an undirected graph $G = (V, E)$ based on the distances in \eqref{eq:rid} using the $\epsilon$-neighborhood criterion, i.e.\ $(i, j) \in E $ iff $d_{ij} < \epsilon$, or $\kappa$-nearest neighbor criterion, i.e.\ $(i, j) \in E$ iff $j$ is one of the $\kappa$ nearest neighbors of $i$. The edge weights $w_{ij}$ are defined using a kernel function on the $\mathcal{G}$-invariant distance $w_{ij} = K_\sigma (d_{ij}) $. For example, the Gaussian
kernel leads to weights of the form 
\begin{equation}
\label{eq:optimal_dist_kernel}
    w_{ij} = K_\sigma(d_{ij}) = \exp\left(-\frac{\min_{g \in \mathcal{G}} \|x_i - g\cdot x_j\|^2}{\sigma} \right).
\end{equation}    
    
The resulting graph is defined on the quotient space $\mathcal{M} := \mathcal{X}/ \mathcal{G}$ and is invariant to the group transformation of the individual data points. Under certain conditions, the quotient space $\mathcal{M}$ is also a smooth manifold. 
We can identify each data point $x_i$ with $v_i \in \mathcal{M}$ and the dimension of $\mathcal{M}$ is lower than the dimension of $\mathcal{X}$. The unitary irreducible representation of the group $g$ is represented by $\rho_k(g)$. If $v_i$ and $v_j$ are close on the manifold, then the representation $\rho_1(g_{ij})$  of the optimal alignment $g_{ij}$ is an approximation of the local parallel transport operator $P_{x_i, x_j} : T_{v_j} \mathcal{M} \mapsto T_{v_i}\mathcal{M}$~\cite{singer2011viewing,SingerWu2012VDM}.  

Take cryo-EM imaging as an example, each image is a tomographic projection of a 3D object at an unknown orientation $x \in \SO(3)$ represented by a $3\times 3$ orthogonal matrix $ R = [R^1, R^2, R^3]$ satisfying $R^\top R = RR^\top = I$ and $\det R = 1$~\cite{singer2011viewing,hadani2011representation,zhao2014rotationally}. The viewing direction of each image can be represented as a point on the unit sphere, denoted by $v$ ($v = R^3$). The first two columns of the orthogonal matrix $R^{1}$ and $R^2$ correspond to the lifted vertical and horizontal axes of the image in the tangent plane $T_v \mS^2$.
Therefore, each image can be represented by a unit tangent vector on the sphere and the base manifold is $\mathcal{M} = \SO(3)/\SO(2) = \mS^2$. 
Images with similar $v$'s are identified as the nearest neighbors and they can be accurately estimated using \eqref{eq:rid} from clean images.  Registering the centered images corresponds to in-plane rotationally aligning the nearest neighbor images according to \eqref{eq:opt_g}. 

In many applications, noise in the observational data affects the estimations of $\mathcal{G}$-invariant distances $d_{ij}$ and optimal alignments $g_{ij}$. This results in shortcut edges in the $\epsilon$-neighborhood graph or $\kappa$-nearest neighbor graph, and connects points on $\mathcal{M}$ where the underlying geodesic distances are large. 



\section{Algorithm}
\label{sec:alg}
To address this issue of shortcut edges induced by noise, we extend VDM using multiple irreducible representations of the compact alignment group. 

\subsection{Affinity and mapping}
We assume the initial graph $G$ is given along with the optimal alignments on the connected edges. For simplicity and because of our interest in cryo-EM image classification, we focus on $\mathcal{G} = \SO(2)$ and we denote the optimal alignment angle by $\alpha_{ij}$. The corresponding frequency-$k$ unitary irreducible representations is $e^{\imath k \alpha_{ij}}$, where $\imath = \sqrt{-1}$. For points that are nearby on $\mathcal{M}$, the alignments should have cycle consistency under the clean case, for example, $k (\alpha_{ij} + \alpha_{jl} + \alpha_{li}) \approx 0 \text{ mod } 2\pi $ for integers $k \in \mathbb{Z}$, if nodes $i$, $j$ and $l$ are true nearest neighbors. To systematically incorporate the alignment information and impose the consistency of alignments, for a given graph $G = (V, E)$, we construct a set of $n\times n$ affinity matrices $W_{k}$, 
\begin{equation}
    W_k(i, j) = 
    \begin{cases} 
    w_{ij}e^{\imath k\alpha_{ij}} &(i, j) \in E, \\ 
    0 &\text{otherwise},
    \end{cases}
\label{eq:Wk}
\end{equation}
where the edge weights according to \eqref{eq:optimal_dist_kernel} are real, $w_{ij} = w_{ji}$ and $\alpha_{ij} = -\alpha_{ji}$ for all $ (i,j) \in E$. At frequency $k$, the weighted degree of node $i$ is:
\begin{equation}
    \mathrm{deg}(i) : = \sum_{j: (i,j) \in E}|W_k(i, j)| = \sum_{j: (i,j) \in E} w_{ij},
\end{equation}
and the degree is identical through all frequencies.
We define a diagonal degree matrix $D$ of size $n\times n$, where the $i^{\text{th}}$ diagonal entry $D(i,i) = \mathrm{deg}(i)$. 



\begin{figure}
\begin{center}
\includegraphics[width = 0.48\textwidth]{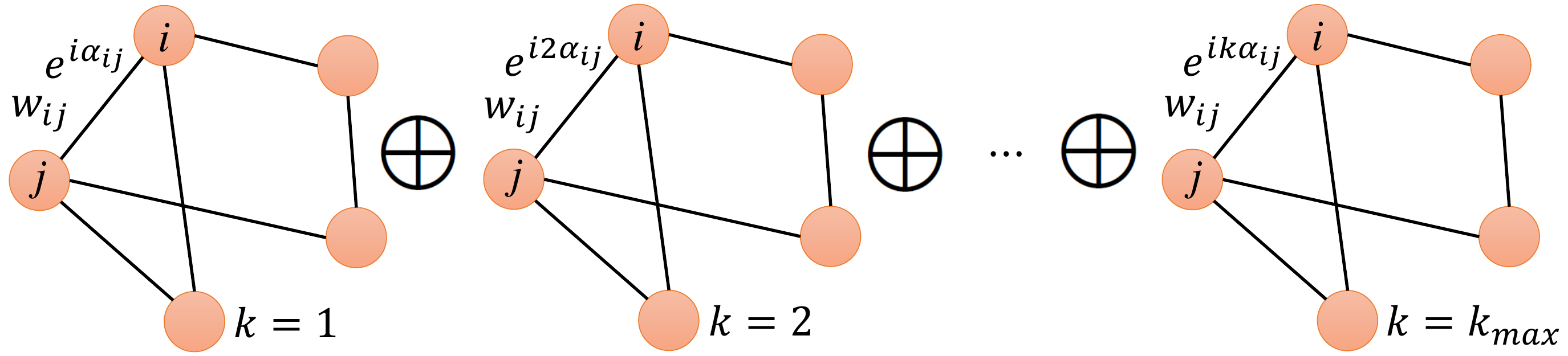}
\end{center}
\vspace{-0.3cm}
\caption{Illustration of multi-frequency edge connection. The $\oplus$-operation denotes concatenation. 
}
\label{fig:graph_illustration}
\vspace{-0.3cm}
\end{figure}

 We construct the normalized matrix $A_k = D^{-1}W_k$ which is applied to complex vectors $z$ of length $n$ and each entry $z(i) \in \mathbb{C}$ can be viewed as a vector in $T \mathcal{M}$. The matrix $A_k$ is an averaging operator for  vector fields, i.e.\ $(A_k z)(i) = \frac{1}{\mathrm{deg}(i)} \sum_{j: (i, j) \in E} w_{ij} e^{\imath k \alpha_{ij}} z(j)$. In our framework, we define affinity between $i$ and $j$ by considering the consistency of the transformations over all paths of length $2t$ that connect $i$ and $j$. In addition, we also consider the consistencies in the transported vectors at $k$ frequency (see Fig.~\ref{fig:graph_illustration}).  
Intuitively, this means $A_k^{2t}(i,j)$ sums the transformations of all length-$2t$ paths from $i$ to $j$, and a large value of $|A_k^{2t}(i,j)|$ indicates not only the strength of connection between $i$ and $j$, but also the level of consistency in the alignment along all connected paths. 

We obtain the affinity of $i$ and $j$ by observing the following decomposition:
\begin{equation}
    A_{k} = D^{-1}W_{k} = D^{-1/2} \underbrace{D^{-1/2}W_{k}D^{-1/2}}_{S_{k}} D^{1/2}.
    \label{eq:A_k}
\end{equation}
Since $S_{k}$ is Hermitian, it has a complete set of real eigenvalues $\lambda_{1}^{(k)}, \lambda_{2}^{(k)},\ldots,\lambda_{n}^{(k)}$ and eigenvectors $u_{1}^{(k)},u_{2}^{(k)},\ldots,u_{n}^{(k)}$, where $\lambda_{1}^{(k)} > \lambda_{2}^{(k)} > \ldots >\lambda_{n}^{(k)}$. We can express $S_{k}^{2t}(i, j)$ in terms of the eigenvalues  and eigenvectors of $S_k$:
\begin{align}
    S_{k}^{2t}(i, j) &= \sum_{l = 1}^{n}\left(\lambda_{l}^{(k)}\right)^{2t} u_{l}^{(k)}(i)\overline{u_{l}^{(k)}(j)}.
\end{align}
Therefore the affinity of $i$ and $j$ at the $k^{\text{th}}$ frequency is given by
\begin{align}
|S_{k}^{2t}(i,j)|^2 & = \sum_{l,r = 1}^{n}\left(\lambda_{l}^{(k)}\lambda_{r}^{(k)}\right)^{2t}u_{l}^{(k)}(i) \overline{u_{r}^{(k)}(i)} \overline{u_{l}^{(k)}(j)} u_{r}^{(k)}(j) \nonumber \\
     &= \left\langle V_t^{(k)}(i), V_t^{(k)}(j)\right\rangle,
     \label{eq:affinity_k_untrun}
\end{align}
which is expressed by an inner product between two vectors $V^{(k)}_t(i), V^{(k)}_t(j) \in \mathbb{C}^{n^2}$ via the mapping $V_{t}^{(k)}$:
\begin{equation}
    V_{t}^{(k)}: i \mapsto \left(\left(\lambda_{l}^{(k)}\lambda_{r}^{(k)}\right)^t \langle u_{l}^{(k)}(i), u_{r}^{(k)}(i) \rangle \right)_{l,r = 1}^n. 
\end{equation}
We call this \textit{frequency-$k$-VDM}. 

\textbf{Truncated mapping:}
Notice the matrices $I + S_k$ and $I - S_k$ are both positive semi-definite (PSD) due to the following property: $\forall z \in \mathbb{C}^n$ we have 
\begin{align}
    &z^{*}(I \pm S_k)z = \\
    &\sum_{(i,j) \in E} w_{ij}\left | \frac{z(i)}{\sqrt{\deg{(i)}}} \pm \frac{ e^{\imath k\alpha_{ij}}z(j)}{\sqrt{\deg(j)}} \right |^2 \geq 0. \nonumber
\end{align}
Therefore all eigenvalues $\{\lambda^{(k)}_i\}_{i = 1}^{n}$ of $S_k$  lie within the interval $[-1,1]$.
Consequently, for large $t$, most  $(\lambda_{l}^{(k)}\lambda_{r}^{(k)})^{2t}$ terms in \eqref{eq:affinity_k_untrun} are close to $0$, and $|S_k^{2t}
(i, j) |^2$  can be well approximated by using only a few of the 
largest eigenvalues and their corresponding eigenvectors. Hence, we truncate the \textit{frequency-$k$-VDM} mapping $V_{t}^{(k)}$ using a cutoff $m_k$ for each frequency $k$: 
\begin{equation}
    \hat{V}_{t}^{(k)}: i \mapsto \left(\left(\lambda_{l}^{(k)}\lambda_{r}^{(k)}\right)^t \langle u_{l}^{(k)}(i), u_{r}^{(k)}(i) \rangle \right)_{l,r = 1}^{m_k}.
    \label{eq:trun_V}
\end{equation}
The affinity of $i$ and $j$ at the frequency $k$ after truncation is given by
\begin{equation}
    |\hat{S}_{k}^{2t}(i, j)|^2 = \left\langle \hat{V}_{t}^{(k)}(i), \hat{V}_{t}^{(k)}(j) \right\rangle \approx |S_k^{2t}(i,j)|^2.
    \label{eq:aff_trun_V_k}
\end{equation}

\textbf{\textit{Remark} 1:}
The truncated mapping not only has the advantage of computational efficiency, but also enhances the robustness to noise since the eigenvectors with smaller eigenvalues are more oscillatory and sensitive to noise. 

\textbf{Multi-frequency mapping:} 
Consider the affinity in \eqref{eq:affinity_k_untrun} for $k = 1,\ldots, k_{\text{max}}$, if $i$ and $j$ are connected by multiple paths with consistent transformations, the affinity $|\hat{S}_{k}^{2t}(i, j)|^2$ should be large for all $k$. Then we can combine multiple representations (i.e., combine multiple $k$) to evaluate the consistencies of the group transformations along connected paths. Therefore, a straightforward way is to 
concatenate the truncated mappings $\hat{V}_{t}^{(k)}$ for all $k = 1,2,\ldots,k_{\text{max}}$ as:
\begin{equation}
    \hat{V}_{t}(i): i \mapsto \left(\hat{V}_{t}^{(1)}(i); \hat{V}_{t}^{(2)}(i); \ldots;\hat{V}_{t}^{(k_{\text{max}})}(i) \right),
\end{equation}
called \textit{multi-frequency vector diffusion maps (MFVDM)}. We define the new affinity of $i$ and $j$ as the inner product of $\hat{V}_{t}(i)$ and $\hat{V}_{t}(j)$:
\begin{align}
|\hat{S}^{2t}(i,j)|^2 &:= \sum_{k = 1}^{k_\text{max}}|\hat{S}_k^{2t}(i,j)|^2 = \sum_{k = 1}^{k_\text{max}}\left\langle \hat{V}_t^{(k)}(i), \hat{V}_t^{(k)}(j)\right\rangle \nonumber \\
      &  =  \left\langle \hat{V}_{t}(i), \hat{V}_{t}(j) \right\rangle.
      \label{eq:affinity_total}        
\end{align}
MFVDM systematically incorporates the cycle consistencies on the geometric graph across multiple irreducible representations of the transformation group elements (in-plane rotational alignments in this case, see Fig.~\ref{fig:graph_illustration}). Using information from multiple irreducible group representations leads to a more robust measure of rotationally invariant similarity.

\textbf{\textit{Remark} 2:}
Empirically, we find the normalized mapping $i \mapsto \frac{\hat{V}_{t}(i)}{\|\hat{V}_{t}(i)\|}$ to be more robust to noise than $\hat{V}_{t}(i)$. A similar phenomenon was discussed in VDM~\cite{SingerWu2012VDM}. The normalized affinity is defined as,
\begin{equation}
     N_t(i, j) = \left\langle \frac{\hat{V}_{t}(i)}{\|\hat{V}_{t}(i)\|},~ \frac{\hat{V}_{t}(j)}{\|\hat{V}_{t}(j)\|}  \right\rangle.
    \label{eq:affinity_norm}
\end{equation}

\textbf{Comparison with DM and VDM:}
Diffusion maps (DM) only consider scalar weights over the edges and the vector diffusion maps (VDM) only take into account consistencies of the transformations along connected edges using only one representation of $\SO(2)$, i.e.\ $e^{\imath \alpha_{ij}}$. In this paper, we generalize VDM and use not only one irreducible representation, i.e.\ $k = 1$, but also higher order $k$ up to $k_{\text{max}}$. 

\subsection{Nearest neighbor search and rotational alignment}
In this section we introduce our method for joint nearest neighbor search and rotational alignment.

\textbf{Nearest neighbor search:} Based on the extended and normalized mapping $i \mapsto \frac{\hat{V}_{t}(i)}{\|\hat{V}_{t}(i)\|}$, we define the multi-frequency vector diffusion distance $d_{\text{MFVDM},t}(i,j)$ between node $i$ and $j$ as
\begin{align}
       & d_{\text{MFVDM},t}^2(i,j) = \left\|\frac{\hat{V}_{t}(i)}{\|\hat{V}_{t}(i)\|} - \frac{\hat{V}_{t}(j)}{\|\hat{V}_{t}(j)\|}\right\|_2^2 \label{eq:dif_distance} \\
       & = 2 - 2\left\langle \frac{\hat{V}_{t}(i)}{\|\hat{V}_{t}(i)\|},~ \frac{\hat{V}_{t}(j)}{\|\hat{V}_{t}(j)\|}  \right\rangle = 2- 2N_t(i,j), \nonumber
\end{align}
which is the Euclidean distance 
between mappings of $i$ and $j$. We define the nearest neighbor for a node $i$ to be the node $j$ with smallest $d_{\text{MFVDM},t}^2(i,j)$. Similarly, for VDM and DM, we define the distances $d_{\text{VDM},t}$ and $d_{\text{DM},t}$,  and perform the nearest neighbor search accordingly. 

\begin{figure}[t!]
    \centering
    \includegraphics[width = 0.25\textwidth]{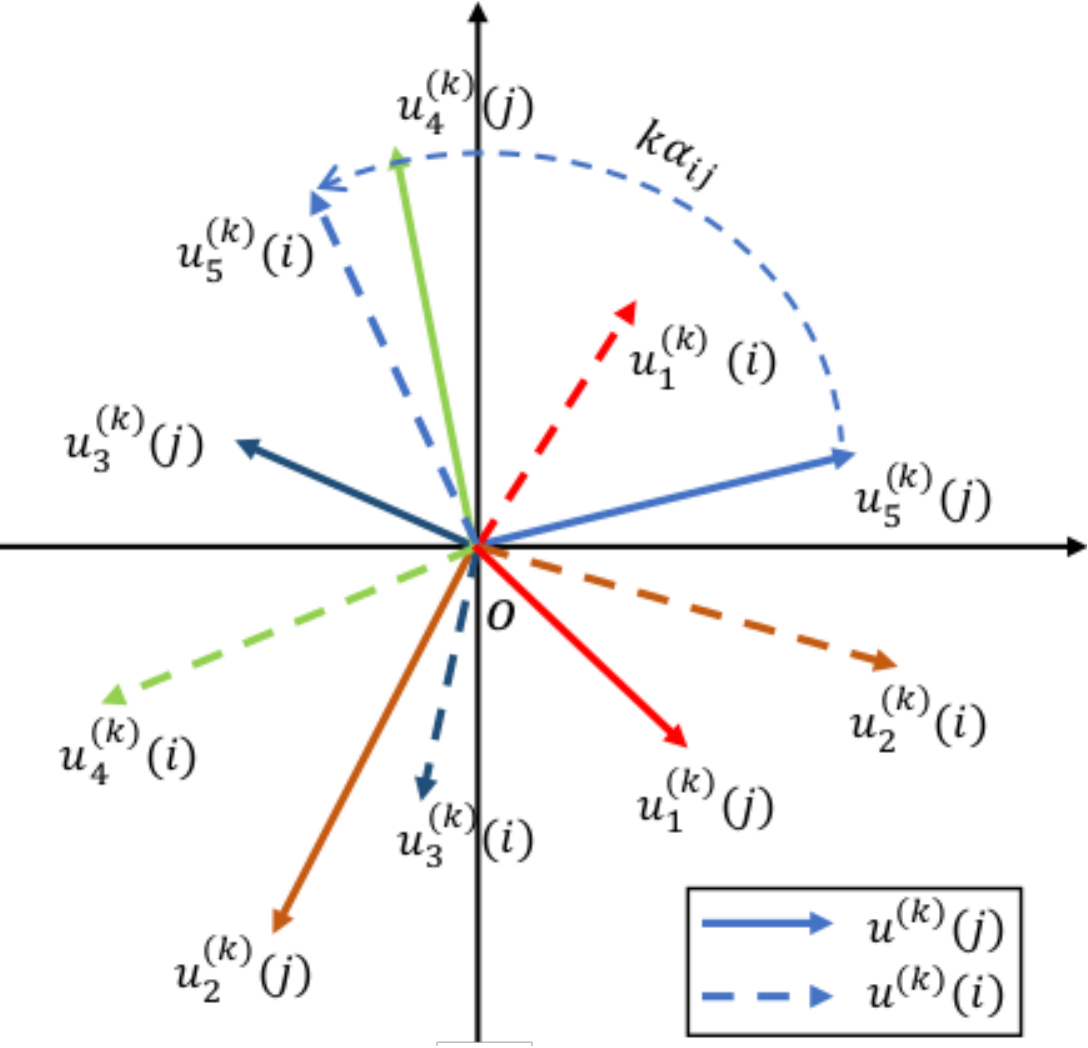}
    \vspace{-0.3cm}
    \caption{Illustration of MFVDM rotational alignment. Solid lines indicate the local frames at node $i$ and dashed lines at node $j$. }
    \label{fig:eVDM_align}
    \vspace{-0.3cm}
\end{figure}

\textbf{Rotational alignment:} 
We notice that the eigenvectors of $S_k$ encode the alignment information between neighboring nodes, as illustrated in Fig.~\ref{fig:eVDM_align}. Assume that two nodes $i$ and $j$ are located at the same base manifold point, for example, the same point on $\mS^2$, but their tangent bundle frames are oriented differently, with an in-plane rotational angle $\alpha_{ij}$. Then the corresponding entries of the eigenvectors are vectors in the complex plane and the following holds, 
\begin{equation}
u^{(k)}_{l}(i) = e^{\imath k \alpha_{ij}} u^{(k)}_l(j), \quad \forall\, l = 1,2,\ldots,n.
\label{eq:sameview}
\end{equation}
When $i$ and $j$ are close but not identical,  \eqref{eq:sameview} holds approximately. Recalling \textit{Remark~1}, due to the existence of noise, for each frequency $k$ we approximate the alignment $e^{\imath k \alpha_{ij}}$ using only top $m_k$  eigenvectors. We then use weighted least squares to estimate $\alpha_{ij}$, which can be written as the following optimization problem:
\begin{align}
\hat{\alpha}_{ij} & = \argmin_\alpha \sum_{k = 1}^{ k_{\text{max}} } \sum_{l = 1}^{m_k} \left(\lambda_l^{(k)}\right)^{2t} \left| u^{(k)}_l(i) - e^{\imath k \alpha} u^{(k)}_l(j) \right|^2 \nonumber \\
& = \argmax_\alpha \sum_{k = 1}^{ k_{\text{max}} } \left( \sum_{l = 1}^{m_k} \left(\lambda_l^{(k)}\right)^{2t} u_l^{(k)}(i) \overline{u_l^{(k)}(j)}\right) e^{-\imath k \alpha} \nonumber\\
&= \argmax_\alpha \sum_{k = 1}^{ k_{\text{max}} }S^{2t}_k(i.j)e^{-ik\alpha}. 
\label{eq:EVDMalign2}
\end{align}

To solve this, we define a sequence $z$ and set $z(k)$ for $k = 1, 2, \dots, k_\text{max}$ to be
\begin{equation}
    z(k) =  S^{2t}_k(i.j) = \sum_{l = 1}^{m_k} \left(\lambda_l^{(k)}\right)^{2t} u_l^{(k)}(i) \overline{u_l^{(k)}(j)}.
    \label{eq:u}
\end{equation}
According to \eqref{eq:u} and \eqref{eq:EVDMalign2}, the alignment angles $\hat{\alpha}_{ij}$ can be efficiently estimated by using an FFT on zero-padded $z$ and identifying its peak. Due to usage of multiple unitary irreducible representations of $\SO(2)$, this approximation is more accurate and robust to noise than VDM. The improvement of the alignment estimation using higher order trigonometric moments is also observed in phase synchronization~\cite{pmlr-v97-gao19f}.




\textbf{Computational complexity: }
Our joint nearest neighbor search and alignment algorithm is summarized in Alg.~\ref{alg:EVDMclass}. The computational complexity is dominated by the eigen-decomposition: Computing the top $m_k$ eigenvectors of the sparse Hermitian matrices $S_k$, for $k = 1, \dots, k_{\text{max}}$ requires $O(\sum_{k = 1}^{k_{\text{max}}} n (m_k^2 + m_k l))$, where $l$ is the average number of non-zero elements in each row of $S_k$ (e.g. number of nearest neighbors). If we assume to use an identical truncation $m$ (i.e., $m_k = m$ for all $k$), and express the above in terms of the mapping dimension $d = k_{\text{max}}m^2$, then the complexity is $O(n (d + l \sqrt{k_{\text{max}} d}))$. For large $d$ and moderate $k_{\text{max}}$, the dominant term is $O(nd)$, therefore MFVDM and VDM ($k_{\text{max}} = 1$) could have similar computational complexity for generating the mapping. Moreover, MFVDM can be faster by parallelizing for each frequency $k$. Next, searching for $\kappa$-nearest neighbors takes $O(n \kappa d \log n)$ flops. The alignment step requires FFT of zero-padded $z$ of length $T$, therefore identifying the alignments takes $O (n \kappa (k_{\text{max}} m + T \log T))$ or $O(n \kappa (\sqrt{k_{\text{max}} d} + T \log T))$. 

\begin{algorithm}[t]
\footnotesize
\SetAlgoLined
\SetAlTitleFnt{\footnotesize}
\SetAlCapNameFnt{\footnotesize}
\SetAlCapFnt{\footnotesize}
\KwIn{Initial noisy nearest neighbor graph $G = (V, E)$ and the corresponding edge weights $w_{ij}e^{\imath k\alpha_{ij}}$ defined on the edges, truncation cutoff $m_k$ for $k = 1, \dots, k_\text{max}$}
\KwOut{$\kappa$-nearest neighbors for each data point and the corresponding alignments $\hat{\alpha}_{ij}$}
\For{$ k = 1, \dots, k_{\mathrm{max}} $}{
~~Construct the normalized affinity matrix $W_k$ and $S_k$ according to \eqref{eq:Wk} and \eqref{eq:A_k}\\
Compute the largest $m_k$ eigenvalues $\lambda^{(k)}_1 \geq \lambda^{(k)}_2, \geq, \dots, \geq \lambda^{(k)}_{m_k}$ of $S_k$ and the corresponding eigenvectors $\{ u^{(k)}_l \}_{l = 1}^{m_k}$ \\
Compute the truncated frequency-$k$ embedding $\hat{V}^{(k)}_{t}$ according to \eqref{eq:trun_V}
}
~~Concatenate the truncated embedding $\{\hat{V}^{(k)}_{t}\}_{k = 1}^{k_{\text{max}}}$, compute the normalized affinity by \eqref{eq:affinity_norm}  \\
Identify $\kappa$ nearest neighbors for each data point \\
Compute $\hat{\alpha}_{ij}$ for nearest neighbor pairs using \eqref{eq:EVDMalign2}.
\caption{Joint nearest neighbor search and alignment}
\label{alg:EVDMclass}
\end{algorithm}



\section{Analysis }
\label{sec:analysis}
We use a probabilistic model to illustrate the noise robustness of our embedding using the top eigenvectors and eigenvalues of $W_k$'s. We start with the clean neighborhood graph, i.e.\ $(i, j) \in E$ if $i$ is among $j$'s $\kappa$-nearest neighbors or $j$ is among $i$'s $\kappa$-nearest neighbors according to the $\mathcal{G}$-invariant distances. 
We construct a noisy graph based on the following process starting from the existing clean graph edges: with probability $p$, the distance $d_{ij}$ is still small and we keep the edge between $i$ and $j$. With probability $1-p$ we remove the edge $(i, j)$ and link $i$ to a random vertex, drawn uniformly at random from the remaining vertices that are not already connected to $i$. We assume that if the link between $i$ and $j$ is a random link, then the optimal alignment $\alpha_{ij}$ is uniformly distributed over $[0, 2\pi)$. Our model assumes that the underlying graph of links between noisy data points is a small-world graph~\cite{watts1998collective} on the manifold, with edges being randomly rewired with probability $1-p$. The alignments take their correct values for true links and random values for the rewired edges. The parameter $p$ controls the signal to noise ratio of the graph connection where $p = 1$ indicates the clean graph.   

The matrix $W_k$ is a random matrix under this model. Since the expected value of the random variable $e^{\imath k \theta}$ vanishes for $\theta \sim \mathrm{Uniform}[0, 2\pi)$, the expected value of the matrix $W_k$ is 
\begin{equation}
\mathbb{E} W_k = p W_k^{\text{clean}},
\end{equation}
where $W_k^{\text{clean}}$ is the clean matrix that corresponds to $p = 1$ obtained in the case that all links and angles are set up correctly. At a single frequency $k$, the matrix $W_k$ can be decomposed into 
\begin{equation}
W_k = p W_k^{\text{clean}} + R_k,  
\end{equation}
where $R_k$ is a random matrix whose elements are independent and identically distributed (i.i.d) zero mean random variables with finite moments, since the elements of $R_k$ are bounded for $1 \leq k \leq k_\text{max}$. 
The top eigenvectors of $W_k$ approximate the top eigenvectors of $W_k^{\text{clean}}$ as long as the 2-norm of $R_k$ is not too large. Various bounds on the spectral norm of random sparse matrices are proven in~\cite{khorunzhy2001sparse,khorunzhiy2003rooted}. This ensures the noise robustness for each frequency-$k$-VDM. Combining an ensemble of classifiers is able to boost the performance~\cite{zhou2012ensemble}.  
Across different frequencies, the entries $R_k$ are dependent through the relations of the irreducible representations. We will provide detailed analysis across frequency channels in the future.


\textbf{Spectral properties for $\boldsymbol{\SO(3)}$:}
Related to the application in cryo-EM image analysis, we assume that the data points $x_i$ are uniformly distributed over $\SO(3)$ according to the Haar measure. The base manifold characterized by the viewing directions $v_i$'s is a unit two sphere $\mS^2$ and the pairwise alignment group is $ \SO(2)$.
Then $e^{\imath k \alpha_{ij}}$ approximates the local parallel transport operator from $T_{v_j} \mS^2$ to $T_{v_i}\mS^2$, whenever $x_i$ and $x_j$ have similar viewing directions $v_i$ and $v_j$ that satisfy $\langle v_i, v_j \rangle  \geq 1 - h$, where $h$ characterizes the size of the small spherical cap of the neighborhood. 
The matrices $W_{k}^{\mathrm{clean}}$ approximate the local parallel transport operators  $P^{(k)}_{h}$, 
which are integral operators over $\SO(3)$. We have the following spectral properties for the integral operators,
\begin{theorem}
\label{thm:eval}
The operator $P^{(k)}_{h}$ has a discrete spectrum $\lambda^{k}_l(h)$, $l \in \mathbb{N}$, with multiplicities equal to $2(l + k) - 1$, for every $h \in (0, 2]$. Moreover, in the regime $h \ll 1$, the eigenvalue $\lambda_l^{(k)}(h)$ has the asymptotic expansion
\begin{equation}
\label{eq:eval}
\lambda^{(k)}_l (h) = \frac{1}{2} h - \frac{ k + (l - 1)( l + 2k)}{8} h^2 + O(h^3).
\end{equation}
\end{theorem}
The proof of Theorem~\ref{thm:eval} is detailed in the Appendix A.1 of~\cite{gao2019repre}. Each eigenvalue $\lambda_l^{(k)}(h)$, as a function of $h$, is a polynomial of degree $ l + k $. This extends Theorem 3 in~\cite{hadani2011representation} to frequencies $k > 1$.   
The multiplicities of the eigenvalues can be seen in the last column of Fig.~\ref{fig:spec_s2} and Fig.~\ref{fig:Cryo_EM_spec}. A direct consequence of Theorem~\ref{thm:eval} is that the top spectral gap of $P^{(k)}_{h}$ for small $h>0$ can be explicitly obtained. 
When $h \ll 1$, the top spectral gap is $ G^{(k)}(h) \approx \frac{ 1 + k }{4} h^2$, which increases with the angular frequency. If we use top $m_k = 2k + 1$ eigenvectors for the frequency-$k$-VDM, then from a perturbation analysis perspective, it is well known (see e.g. \cite{RCY2011,EBW2017,FWZ2018} and the references therein) that the stability of the eigenmaps essentially depends on the top spectral gap. Therefore, we are able to jointly achieve more robust embedding and nearest neighbor search under high level of noise or a large number of outliers. Moreover, we are not restricted to use only top $2k + 1$ eigenvectors and incorporating more eigenvectors can improve the results~\cite{singer2011viewing}. 



\section{Experiments}
\label{sec:exp}


\subsection{Synthetic examples on 2 dimensional sphere and torus}
We test MFVDM on two synthetic examples: 2-D sphere $\mS^2$ and torus $\mathrm{T}^2$. For the first example, we simulate $n = 10^4$ points $x_i$ uniformly distributed over $\SO(3)$ according to the Haar measure. Each $x_i$ can be represented by a $3\times 3$ orthogonal matrix $R_i$ whose determinant is equal to 1. The third column of the rotation matrices $R_i$ (denoted as $v_i$) forms a point on the manifold $\mS^2$, 
\begin{equation}
    \mS^2 = \{v \in \mathbb{R}^3 : \left\| v \right\| = 1\}.
\end{equation}
The pairwise alignment $\alpha_{ij}$ is computed based on~\eqref{eq:opt_g}. 
The hairy ball theorem~\cite{milnor1978analytic} says that a continuous tangent vector field to the two dimensional sphere must vanish at some points on the sphere, therefore, we cannot identify $\alpha_i \in [0, 2\pi)$ for $i = 1, \dots, n$, such that $\alpha_{ij} = \alpha_i - \alpha_j$, for all $i$ and $j$. As a result, we cannot globally align the tangent vectors. For the torus, we sample $n = 10^4$ points uniformly distributed on the manifold, which are embedded in three dimensional space according to,
\begin{equation}
    \mathrm{T}^2 = 
    \begin{cases}
        x = \left(R + r\cos{u}\right)\cos{v},\\
        y = \left(R + r\cos{u}\right)\sin{v},\\
        z = r\sin{u},
    \end{cases}
\end{equation}
where $R = 1$, $r = 0.2$ and $(u,v) \in [0,2\pi)\cup[0,2\pi)$, and for each node $i$ we assign an angle $\alpha_i$ that is uniformly distributed in $[0, 2\pi)$, due to the existence of a continuous vector field, we set the pairwise alignment $\alpha_{ij} = \alpha_i - \alpha_j$. For both examples, we connect each node with its top 150 nearest neighbors based on their geodesic distances on the base manifold, then noise is added on edges following the random graph model described in Sec.~\ref{sec:analysis} with parameter $p$. Finally, we build the affinity matrix $W_{k}$ by setting weights $w_{ij} \equiv 1$ $\forall (i,j) \in E$, with $k = 1,2,\ldots, k_{\text{max}}$.

\textbf{Parameter setting:} 
For MFVDM, we set the maximum frequency $k_{\text{max}} = 50$ and for each $k$, we select top $m_k = 50$ eigenvectors. For VDM and DM, we set the number of eigenvectors to be $m = 50$. In addition, 
we set random walk step size $t = 1$. 

\begin{figure}[!t]
	\vspace{-0.15cm}
	
	\centering
	\includegraphics[width= 0.90\linewidth]{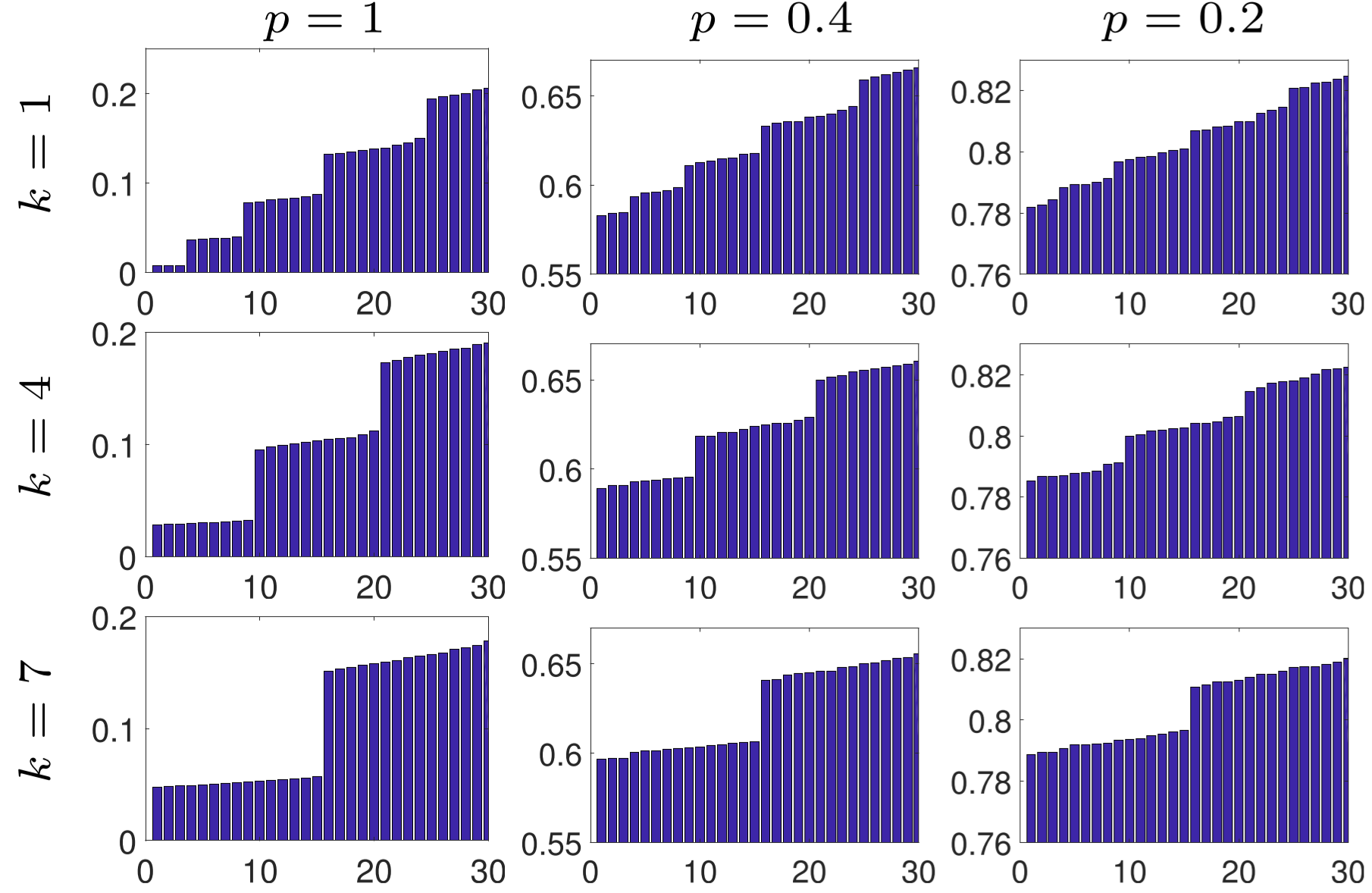}\\[-10pt]
	
	\caption{$\mS^2$ case: Bar plots of the 30 smallest eigenvalues $1 - \lambda^{(k)}$ of the graph connection Laplacian $I - S_k$ on $\mS^2$ for different $p$'s and $k$'s.
	}
	\label{fig:spec_s2}
\end{figure}

\textbf{Spectral property on $\mS^2$:} We numerically verify the spectrum of graph connection Laplacian $I - S_{k}$ on $\mS^2$ for different $k$ and random rewiring parameter $p$. Smaller $p$ indicates more edges are corrupted by noise. 
Fig.~\ref{fig:spec_s2} shows that the multiplicities of $S_{k}$ (normalized $W_k$ matrix) agree with Theorem~\ref{thm:eval}. The spectral gaps persist even when 80\% of the edges are corrupted (see the right column of Fig.~\ref{fig:spec_s2}). 

\begin{figure}[!t]
	\vspace{-0.45cm}
	\hspace{0.3cm}
	\includegraphics[width= 0.88\linewidth]{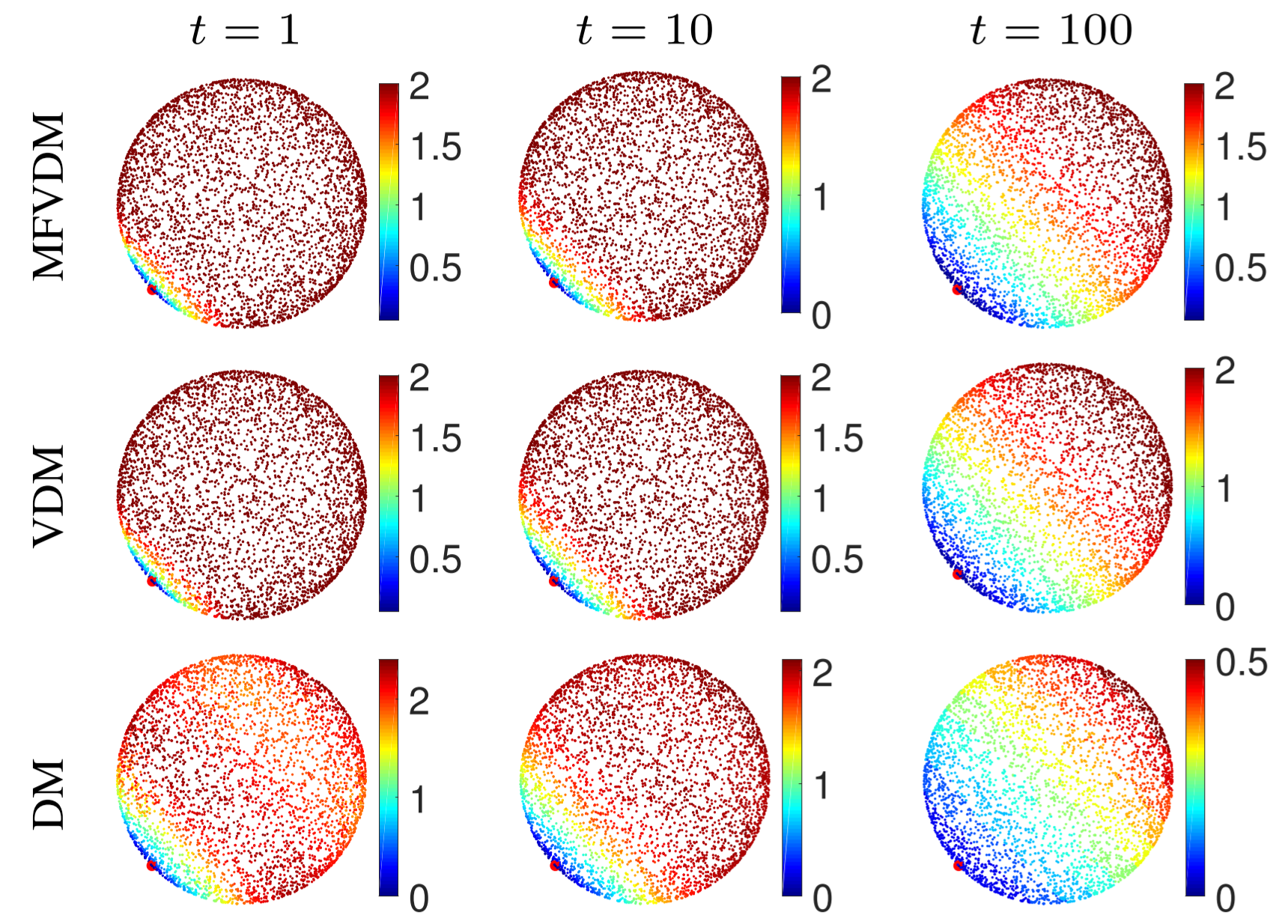}\\
	
	\vspace{-0.8cm}
	\caption{$\mS^2$ case: The normalized $d_{\text{MFVDM}, t}$, $d_{\text{VDM}, t}$, and $d_{\text{DM}, t}$ between a reference point (marked in red) and other points, with $t = 1, 10, \text{and } 100$, $p = 1$.}
	\label{fig:diffusion_dis_sphere}
\end{figure}

\begin{figure}[t]
	\centering
	\vspace{-0.2cm}
	\hspace{0.02cm}
	\includegraphics[width= 0.88\linewidth]{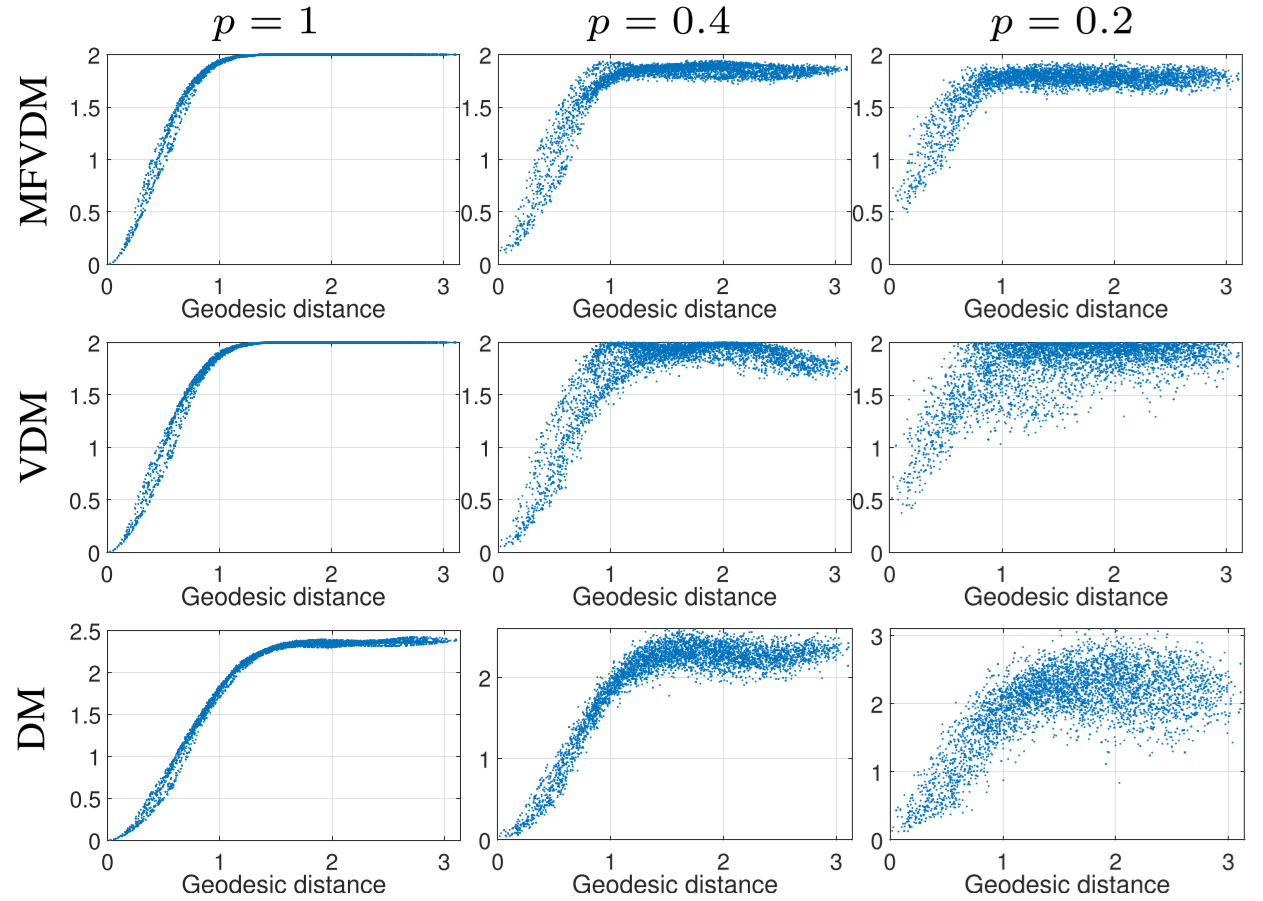}\\[-10pt]
	\caption{$\mS^2$ case: Scatter plots comparing the normalized $d_{\text{MFVDM}, t}$, $d_{\text{VDM}, t}$, and $d_{\text{DM}, t}$ at $p = 0.2, 0.4,$ and 1.}
	\label{fig:scatter_sphere}
\end{figure}

\textbf{Multi-frequency vector diffusion distances on $\mS^2$:} 
Based on \eqref{eq:dif_distance}, Fig.~\ref{fig:diffusion_dis_sphere} displays the normalized and truncated multi-frequency vector diffusion distances $d_{\text{MFVDM},t}^2(i,j)$, vector diffusion distances $d_{\text{VDM},t}^2(i,j)$, and diffusion distances $d_{\text{DM},t}^2(i,j)$ between a reference point (marked in red) and others, on $\mS^2$ at $p = 1$ (clean graph). Moreover, we increase the diffusion step size $t$ from $t = 1$ to $t = 10\text{ and }100$. In this clean case, all three distances are highly correlated to the geodesic distance. Specifically, MFVDM and VDM perform similarly. 

To demonstrate the robustness to noise of $d_{\text{MFVDM},t}$, we compare $d_{\text{MFVDM},t}$, $d_{\text{VDM},t}$, and $d_{\text{DM},t}$ against the geodesic distance on $\mS^2$ in Fig.~\ref{fig:scatter_sphere} 
at different noise levels. When $p = 1$, all the distances are highly correlated with the geodesic distance, 
e.g., small $d_{\text{MFVDM},t}$, $d_{\text{VDM},t}$, and $d_{\text{DM},t}$ all correspond to small geodesic distance.
However at high noise level as $p = 0.4$ or $0.2$, both $d_{\text{VDM},t}$ and $d_{\text{DM},t}$ become more scattered, while $d_{\text{MFVDM},t}$ remains correlated with the geodesic distance. Here the random walk steps $t = 10$ and the results are similar for $t = 1$ or $100$. 

\textbf{Nearest neighbor search and rotational alignment:}
We test the nearest neighbor search (NN search) and rotational alignment results on both sphere and torus, with different noise levels $p$. As mentioned, one advantage of MFVDM is its robustness to noise. Even at a high noise level, the true affinity between nearest neighbors can still be preserved. In our experiments, for each node we identify its $\kappa = 50$ nearest neighbors.

\begin{figure}[t]
	\centering
	\vspace{-0.2cm}
	\includegraphics[width= 1.0\linewidth]{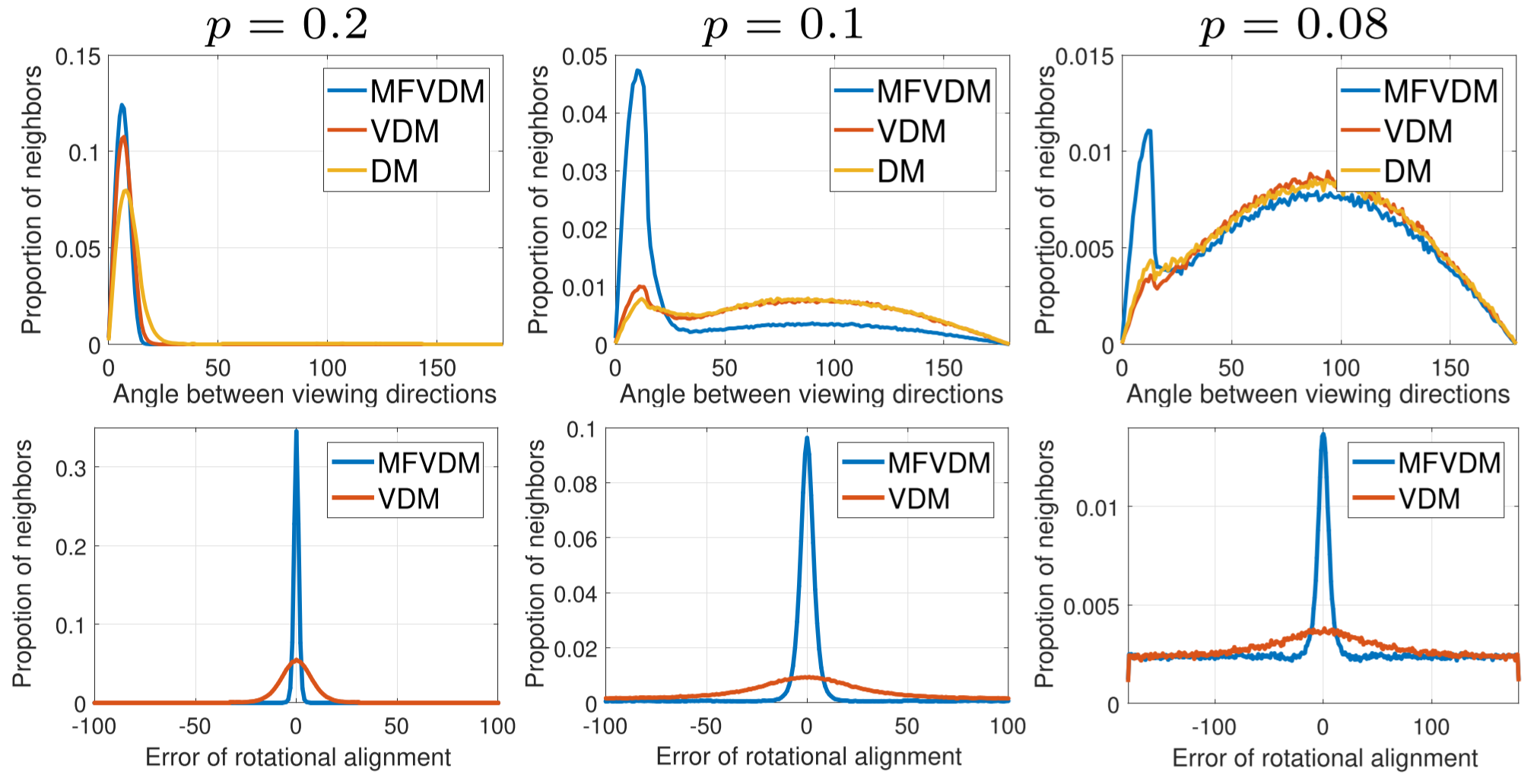}\\[-10pt]
	\caption{$\mS^2$ case: \textit{Top}: histograms of the viewing direction difference between nearest neighbors found by MFVDM, VDM and DM; \textit{Bottom}: the accuracy of the rotational alignment estimated by MFVDM and VDM. 
	}
	\label{fig:class_rot_sphere}
\end{figure}

\begin{figure}[t]
	\centering
	\vspace{-0.35cm}
	\includegraphics[width= 1.0\linewidth]{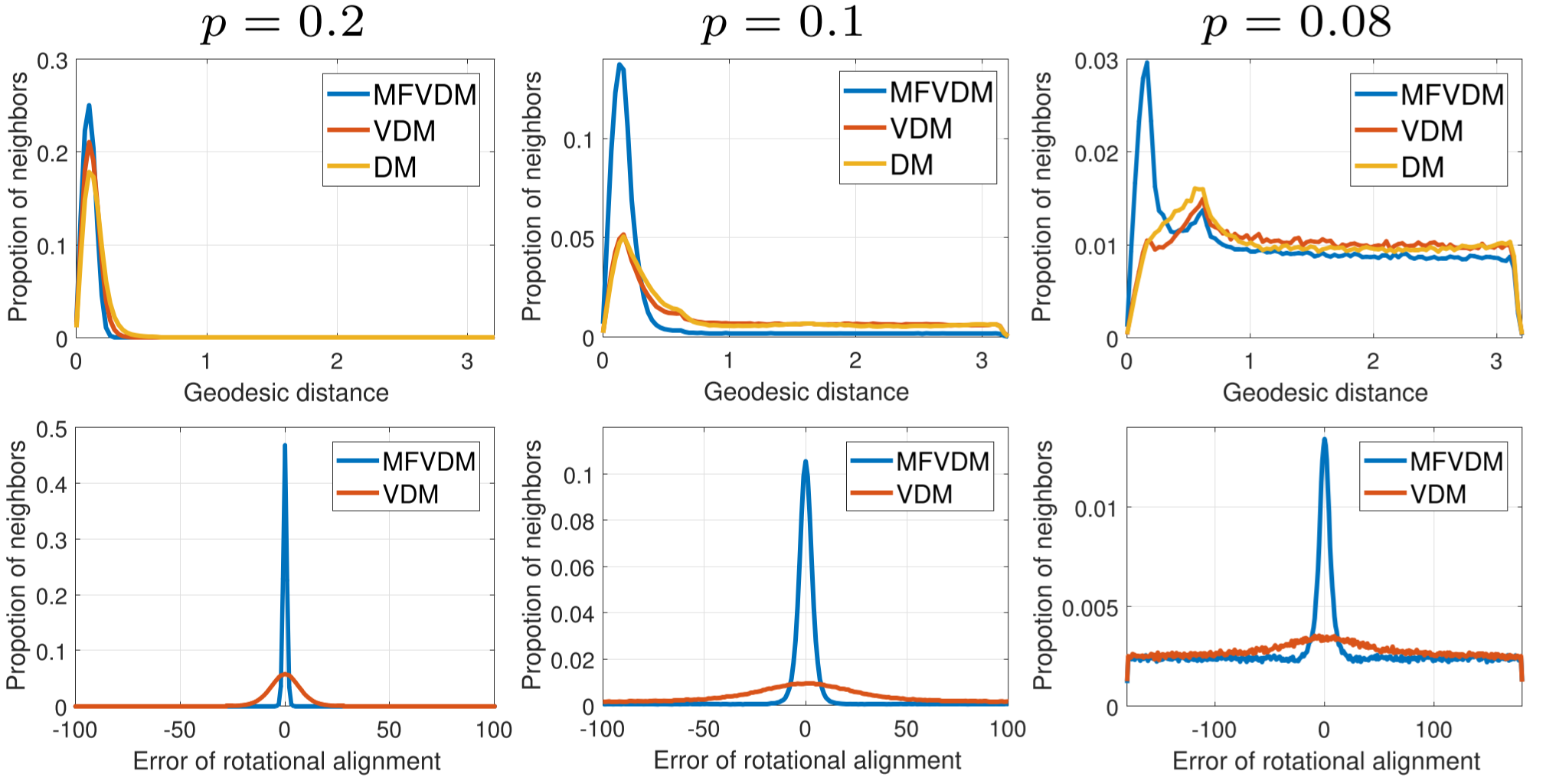}\\[-10pt]
	\caption{$\mathrm{T}^2$ case: \textit{Top}: histograms of the geodesic distances between nearest neighbors identified by MFVDM, VDM and DM; \textit{Bottom}: the accuracy of the rotational alignment estimated by MFVDM and VDM. }
	\label{fig:class_rot_torus}
\end{figure}

We evaluate the NN search by the geodesic distance between each node and its nearest neighbors. A better method should find more neighbors with geodesic distance close to 0. In the top rows of Fig.~\ref{fig:class_rot_sphere} and Fig.~\ref{fig:class_rot_torus} we show the histograms of such geodesic distance. Note that in the low noise regime ($p \geq 0.2$), MFVDM, VDM and DM all perform well and MFVDM is slightly better. When the noise level increases to $p = 0.1$, both VDM and DM have poor result while MFVDM still works well. 
These comparisons show MFVDM, which benefits from multiple irreducible representations, is very robust to noise. 

We evaluate the rotational alignment estimation by computing the alignment errors $\alpha_{ij} - \hat{\alpha}_{ij}$ for all pairs of nearest neighbors $(i,j)$, where $\alpha_{ij}$ is the ground truth and $\hat{\alpha}_{ij}$ is the estimation. In the bottom rows of Fig.~\ref{fig:class_rot_sphere} and Fig.~\ref{fig:class_rot_torus}, we show the histograms of such alignment errors. The results demonstrate that for a wide range of $p$, i.e., $p \geq 0.1$, the MFVDM alignment errors are closer to $0$ than the baseline VDM. 
At $p = 0.08$, the VDM errors disperse between 0 to 180 degrees, whereas a large number of the alignment errors of MFVDM are still close to 0.


\begin{figure}[t!]
    \centering
    \vspace{-0.3cm}
   	\includegraphics[width= 1.0\linewidth]{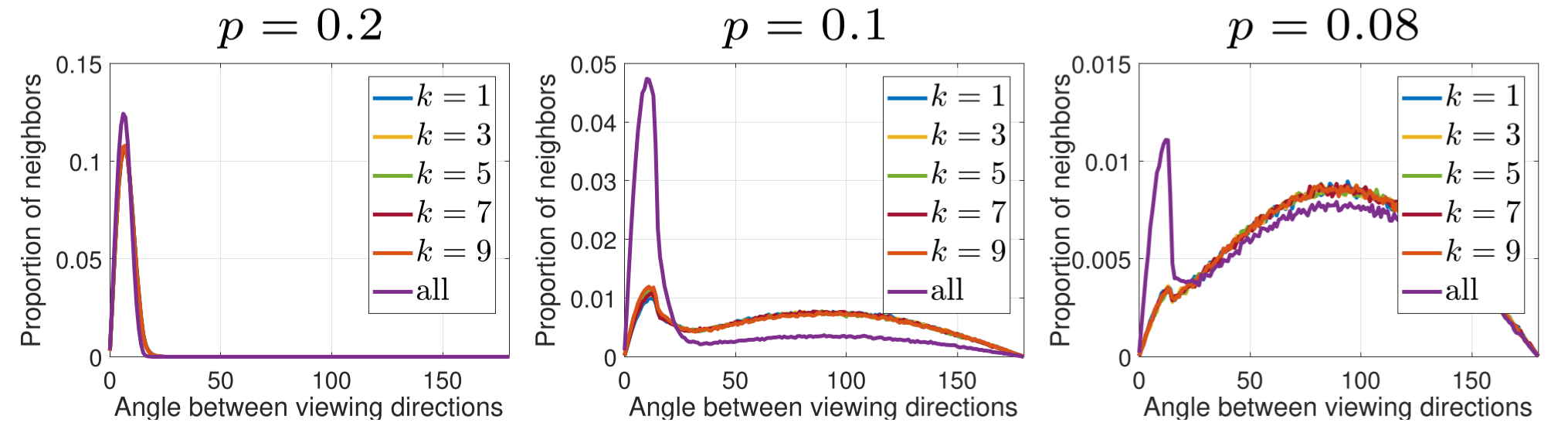}\\[-10pt]
    \caption{$\mS^2$ case: \textit{Weak classifier versus strong classifier}: histograms of the angles between nearest neighbors found by using single frequency-$k$-VDM (weak classifier) and MFVDM (all) with $k = 1,\ldots, k_{\text{max}}$ (strong classifier, shown as `all'). Here $k_{\text{max}} = 10$. }
    \label{fig:weak_vs_strong}
\end{figure}

At each frequency $k$, we individually perform NN search based on frequency-$k$-VDM and the corresponding affinity in~\eqref{eq:aff_trun_V_k}. For the $\mS^2$ example, we find that all single frequency mappings achieve similar accuracies when $m_k$'s are identical (see Fig.~\ref{fig:weak_vs_strong}). MFVDM combines those weak single frequency classifiers into a strong classifier to boost the accuracy of nearest neighbor search


\begin{figure}[t!]
    \vspace{-0.2cm}
    \centering
    \subfloat{\includegraphics[width = 0.45\textwidth]{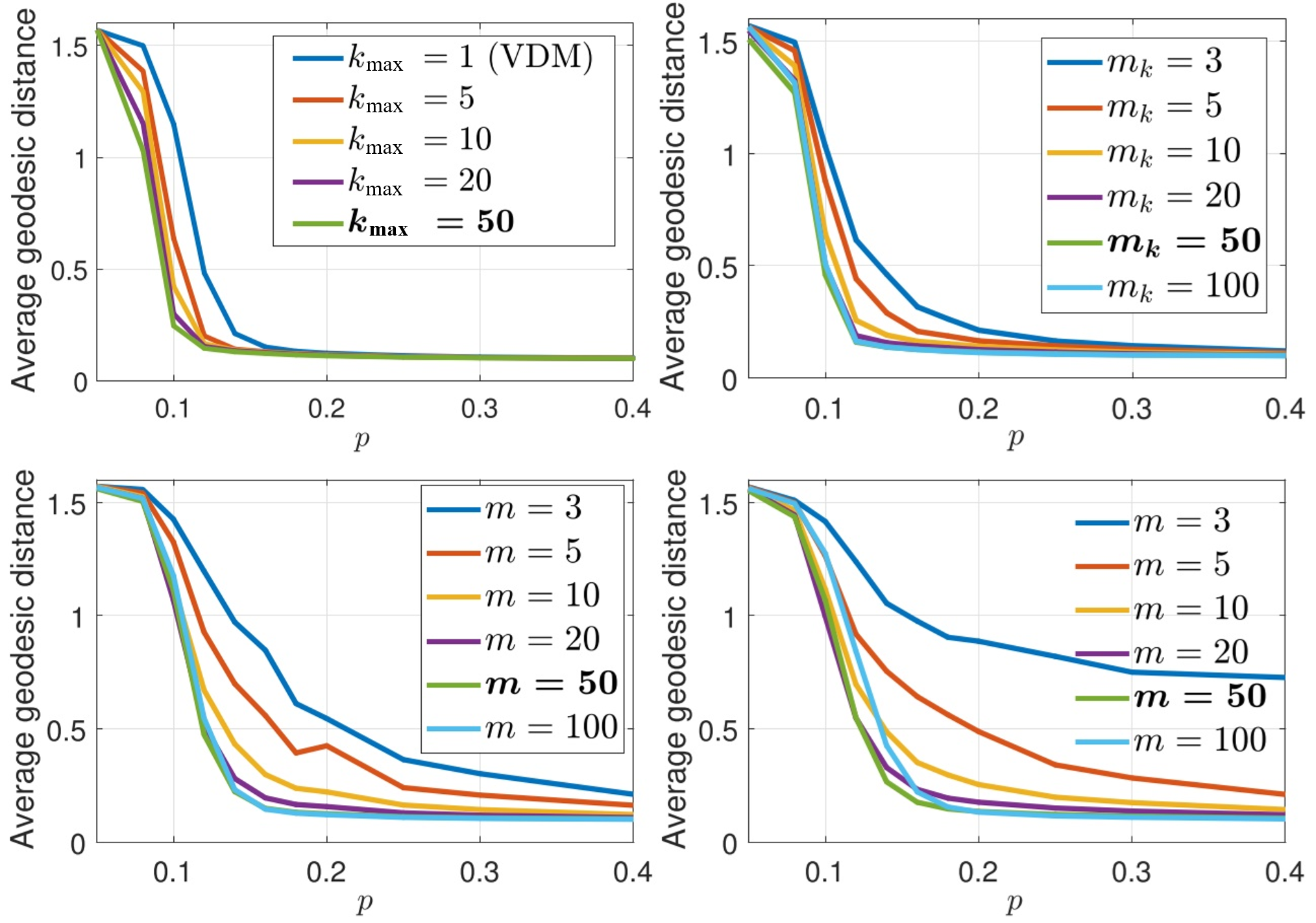}}
    \vspace{-0.4cm}
    \caption{$\mS^2$ case: Nearest neighbor search by MFVDM, VDM and DM under varying parameters: maximum frequency $k_\m$ and the number of eigenvectors $m_k$. \textit{Upper left}: MFVDM with varying $k_{\text{max}}$ and $m_k = 50$; \textit{Upper right}: MFVDM with varying $m_k$ and $k_{\text{max}} = 10$; \textit{Lower left}: VDM with varying $m$ ($k_{\text{max}} = 1$); \textit{Lower right}: DM with varying $m$. Horizontal axis: the value of the parameter $p$ in the random graph model, lower $p$ means larger number of outliers in the edge connections. Vertical axis: the average geodesic distances of nearest neighbors pairs (lower is better). }
    \label{fig:parameter}
\end{figure}

\textbf{Choice of parameters: } 
The performance of MFVDM depends on two parameters: the maximum frequency cutoff $k_{\text{max}}$ and the number of top eigenvectors $m_k$. We assume that $m_k$'s are the same for all frequencies, that is $m_1 = m_2 = \dots = m_{k_\m} = m_k$. 
In the top row of Fig.~\ref{fig:parameter}, we show the average geodesic distances between the nearest neighbor pairs identified by MFVDM, with different values of $k_{\text{max}}$ and $m_k$. First, we fix $m_k = 50$ and vary $k_\m$. The performance of MFVDM improves with increasing $k_\m$ and plateaus when $k_\m$ approaches 50 (see the upper left panel of Fig.~\ref{fig:parameter}).  
Then we fix  $k_\m = 10$ and vary $m_k$. The upper right panel of Fig.~\ref{fig:parameter} shows that choosing $m_k = 50$ achieves the best performance. Using a larger number of eigenvectors, i.e. $m_k = 100$, does not lead to higher accuracy in nearest neighbor search, because the eigenvectors of $S_k$ with small eigenvalues are more sensitive to noise and including them will reduce the robustness to noise of the mappings. 
In addition, we evaluate the performance of VDM and DM under varying number of eigenvectors $m$ in the bottom row of Fig.~\ref{fig:parameter}. VDM and DM also achieve the best performance at $m = 50$. Comparing the upper left and lower left panels of Fig.~\ref{fig:parameter}, we find that MFVDM greatly improves the nearest neighbor search accuracy of VDM when 90\% of the true edges are rewired. Note that the solid blue line in the upper left panel of Fig.~\ref{fig:parameter} corresponds to the best performance curve in the lower left panel of Fig.~\ref{fig:parameter} (green line with $m = 50$).  

\subsection{Application: Cryo-EM 2-D image analysis}
MFVDM is motivated by the cryo-EM 2-D class averaging problem. In the experiments, protein samples are frozen in a very thin ice layer. Each image is a tomographic projection of the protein density map at an unknown random orientation. It is associated with a $3 \times 3$ rotation matrix $R_i$, where the third column of $R_i$ indicates the projection direction $v_i$, which can be realized by a point on $\mS^2$. Projection images $I_i$ and $I_j$ that share the same views look the same up to some in-plane rotation. 
The goal is to identify images with similar views, then perform local rotational alignment and averaging to denoise the image. Therefore, MFVDM is suitable to perform the nearest neighbor search and rotational alignment estimation. 

\begin{figure}[t!]
\captionsetup[subfigure]{labelformat=empty}
\vspace{-0.2cm}
    \centering
    \setlength\tabcolsep{2pt}
    \begin{tabular}{p{0.11\textwidth}<{\centering} p{0.11\textwidth}<{\centering} p{0.11\textwidth}<{\centering}}
         \scriptsize{Reference volume}
         &\scriptsize{Clean projection} 
         &\scriptsize{SNR = 0.05}\\
         \includegraphics[width = 0.105\textwidth]{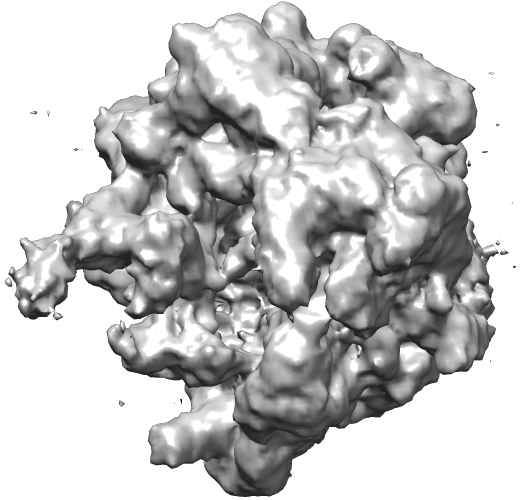}
         &\includegraphics[width = 0.11\textwidth]{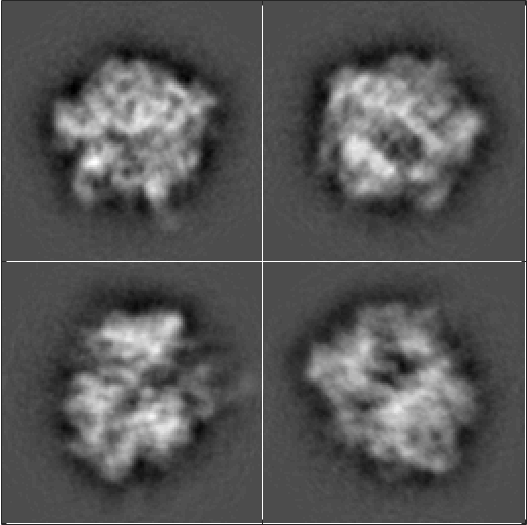}
         &\includegraphics[width = 0.11\textwidth]{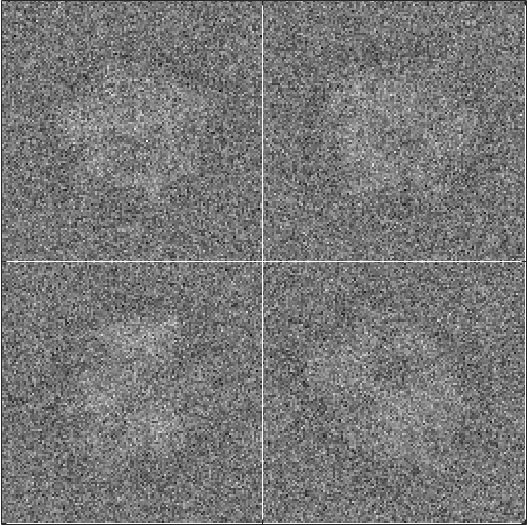}\\
    \end{tabular}
    \vspace{-0.3cm}
    \caption{Cryo-EM 2-D image analysis: \textit{Left}: Reference volume of 70S ribosome; \textit{Mid}: Clean projection images; \textit{Right}: Noisy projection images at SNR$ = 0.05$.}
    \label{fig:Cryo_EM_proj}
\end{figure}


In our experiment, we simulate $n = 10^4$ projection images from a 3-D electron density map of the 70S ribosome (see Fig.~\ref{fig:Cryo_EM_proj}), the orientations for the projection images are uniformly distributed over $\SO(3)$ and the images are contaminated by additive white Gaussian noise at signal-to-noise ratio (SNR) equal to 0.05. Note that such high noise level is commonly observed in real experiments. In Fig.~\ref{fig:Cryo_EM_proj}, we display samples of such clean and noisy images. 
We use fast steerable PCA (sPCA)~\cite{zhao2016fast} and rotationally invariant features~\cite{zhao2014rotationally} to initially identify the images of similar views and the in-plane rotational alignment angles according to~\cite{zhao2014rotationally}. Then we take the initial graph structure and the estimated optimal alignments as the input of Alg.~\ref{alg:EVDMclass}. 

\begin{figure}[t!]
    \centering
    \vspace{-0.4cm}
    \includegraphics[width = 1.0\linewidth]{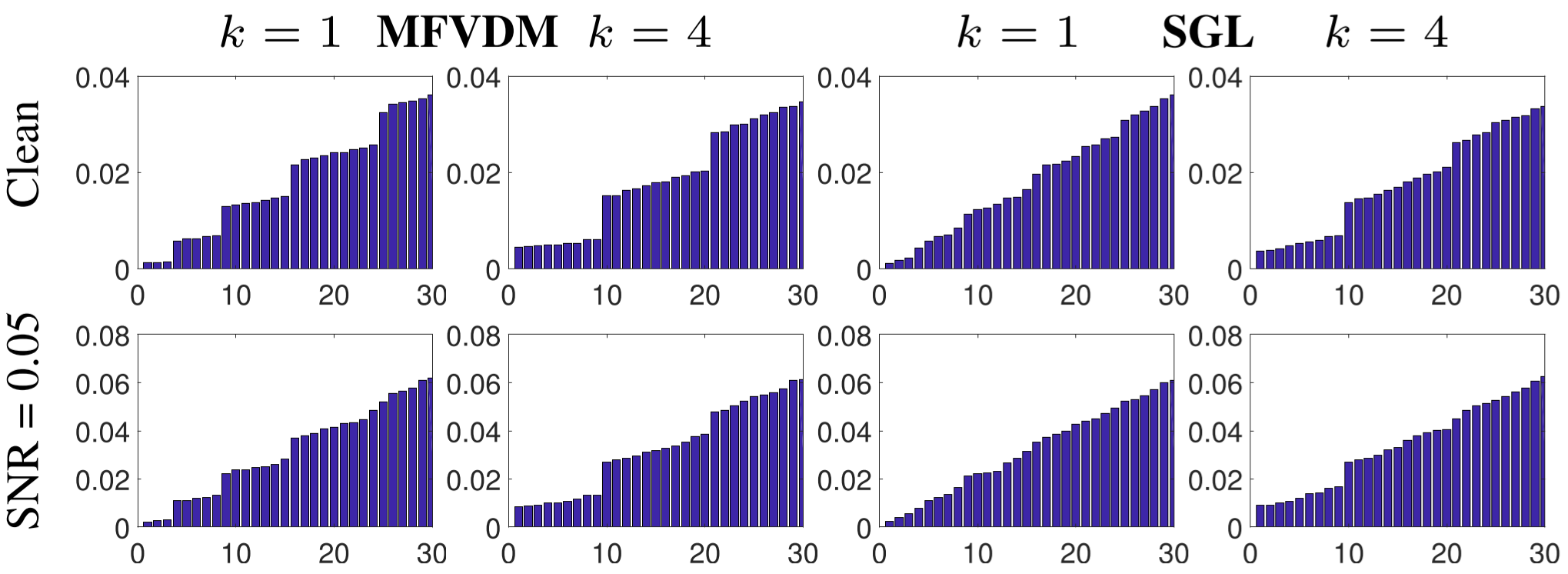}\\[-10pt]
    \caption{Bar plots of the 30 smallest eigenvalues of the graph connection Laplacian $I - S_k$ that is built upon the initial NN search and alignment results on cryo-EM images (MFVDM) and the corresponding eigenvalues of the steerable graph Laplacian (SGL).}
    \vspace{-0.2cm}
    \label{fig:Cryo_EM_spec}
\end{figure}

\begin{figure}[t]
    \centering
    \vspace{-0.2cm}
    \includegraphics[width = 1.0\linewidth]{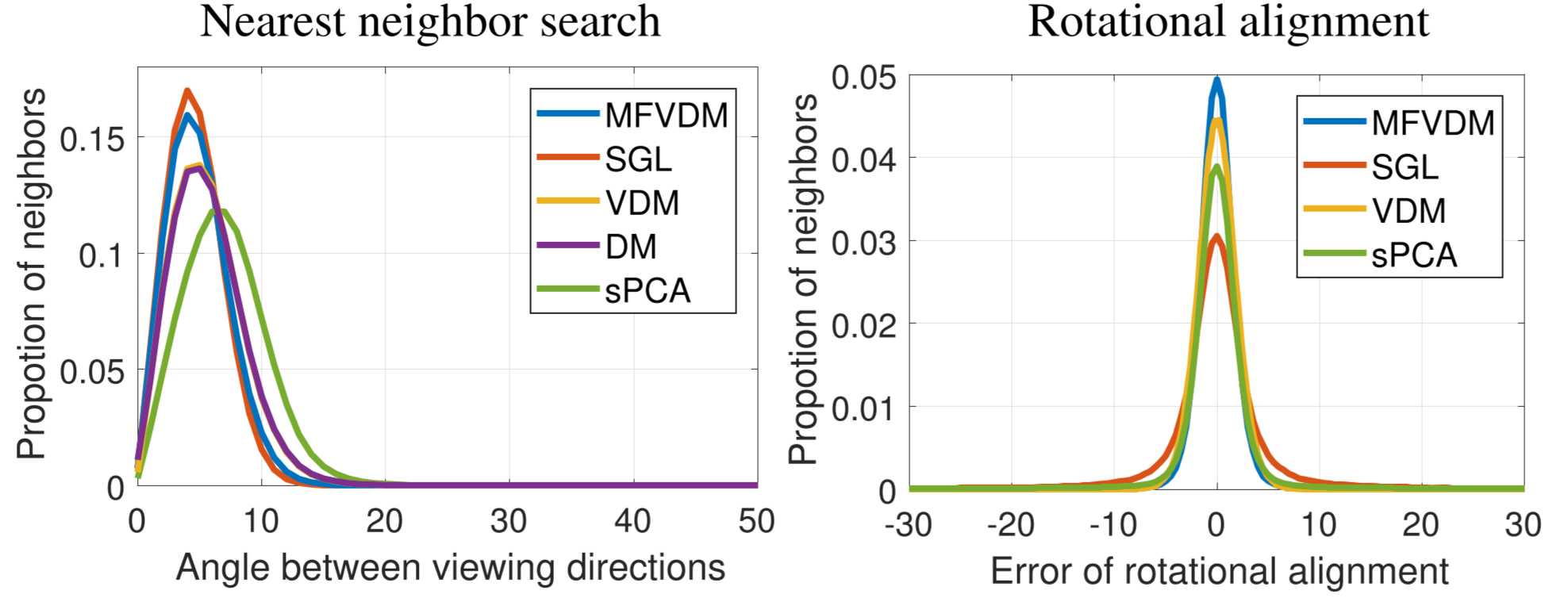}\\[-10pt]
    \caption{Nearest neighbor search and rotational alignment estimation for simulated cryo-EM images of 70S ribosome at SNR = 0.05. \textit{Left}: the distributions of the viewing angles between the estimated nearest neighbors. 
    \textit{Right}: the estimation errors of the optimal in-plane rotational alignments.}
     \vspace{-0.4cm}
    \label{fig:Cryo_EM_class_rot}
\end{figure}

In addition to DM, VDM, and MFVDM, we apply another type of kernel introduced in steerable graph Laplacian (SGL)~\cite{landa2018steerable}, which is defined on image pairs considering all possible rotational alignments, to the image datasets. 
In Fig.~\ref{fig:Cryo_EM_spec}, we present 30 smallest eigenvalues of the graph connection Laplacian $I-S_k$. The spectral gaps are more prominent for both clean and noisy images with MFVDM. 
We set $t= 10$, $k_{\text{max}} = 10$, $m_k = 10$, and $m = 10$ for MFVDM, VDM, and DM respectively.  
Although using SGL kernel achieves slightly better NN search, its performance on alignment estimation is worse than MFVDM (see Fig.~\ref{fig:Cryo_EM_class_rot}). 

\section{Discussion}
\label{sec:dis}
In the current probabilistic model, we only consider independent edge noise, i.e., the entries in $R_k$ for a fixed $k$ are independent. This does not cover the measurement scenarios in some applications. For example, in cryo-EM 2-D image analysis, each image is corrupted by independent noise. Therefore, the entries in $R_k$ become dependent since the edge connections and alignments are affected by the noise in each image node. Empirically, our new algorithm is still applicable and results in the improved nearest neighbor search and rotational alignment estimation compared to the state-of-the-art VDM. We leave the analysis of node level noise to future work. In addition, there can be other approaches to define the multi-frequency mapping, such as weighted average among different frequencies or majority voting. We will explore other ways to integrate multi-frequency information in the future.

The current analysis focuses on data points that are uniformly distributed on the manifold. For non-uniformly distributed data points, different normalization techniques introduced in DM~\cite{coifman2006diffusion,zelnik2005self} are needed to compensate for the non-uniform sampling density.

Since our framework is motivated by the cryo-EM nearest neighbor image search and alignment, we have so far only considered the compact manifold $\mathcal{M}$ where the intrinsic dimension is $2$ and the local parallel transport operator can be well approximated by the in-plane rotational alignment of the images or the alignment of the local tangent bundles as discussed in VDM~\cite{SingerWu2012VDM}. In the future, we will extend the current algorithm to manifolds with higher intrinsic dimension and other compact group alignments $g \in \mathcal{G}$ with their corresponding irreducible representations $\rho_k(g)$, for example, the symmetric group which is widely used in computer vision~\cite{bajaj2018smac}. 

\vspace{-0.2cm}
\section{Conclusion}
\vspace{-0.2cm}
\label{sec:con}
In this paper, we have introduced MFVDM for joint nearest neighbor search and rotational alignment estimation. The key idea is to extend VDM using multiple irreducible representations of the compact Lie group. 
Enforcing the consistency of the transformations at different frequencies allows us to achieve better nearest neighbor identification and accurately estimate the alignments between the updated nearest neighbor pairs. 
The approach is based on spectral decomposition of multiple kernel matrices. We use the random matrix theory and the rationale of ensemble methods to justify the robustness of MFVDM. 
Experimental results show efficacy of our approach compared to the state-of-the-art methods. 
This general framework can be applied to many other problems, such as joint synchronization and clustering~\cite{gao2016geometry} and multi-frame alignment in computer vision.



\bibliography{./ref_AISTATS}
\bibliographystyle{icml2019}

\end{spacing}
\end{document}